\documentclass[11pt]{article}
\usepackage{coling2020}
\usepackage{times}
\usepackage{url}
\usepackage{latexsym}
\usepackage{graphicx}
\usepackage{amsfonts}
\usepackage{amsmath}
\usepackage{url}
\usepackage{multirow}
\usepackage{xcolor}
\usepackage{booktabs}

\newcommand{\draftAZ}[1]{{#1}}
\newcommand{\draftMT}[1]{{#1}}

\def\Bx{\mbox{\boldmath $x$}}
\def\By{\mbox{\boldmath $y$}}

\def\Bh{\mbox{\boldmath $h$}}
\def\Be{\mbox{\boldmath $e$}}
\def\Bs{\mbox{\boldmath $s$}}
\def\Ba{\mbox{\boldmath $a$}}

\def\Bc{\mbox{\boldmath $c$}}

\def\Bo{\mbox{\boldmath $o$}}
\def\Bf{\mbox{\boldmath $f$}}
\def\Bk{\mbox{\boldmath $k$}}

\colingfinalcopy % Uncomment this line for the final submission

\title{Vulgaris: Analysis of a Corpus for Middle-Age Varieties of Italian Language}

\author{Andrea Zugarini$^1,^2$ \and Matteo Tiezzi$^2$ \and Marco Maggini$^2$
\\ $^1$DINFO, University of Florence \\ $^2$DIISM, University of Siena\\
{\tt andrea.zugarini@unifi.it, \{mtiezzi,maggini\}@diism.unisi.it} }

\date{}

\begin{document}
\maketitle
\begin{abstract}
  Italian is a Romance language that has its roots in Vulgar Latin. The birth of the modern Italian started in Tuscany around the 14th century, and it is mainly attributed to the works of Dante Alighieri, Francesco Petrarca and Giovanni Boccaccio, who are among the most acclaimed authors of the medieval age in Tuscany. However, Italy has been characterized by a high variety of dialects, which are often loosely related to each other, due to the past fragmentation of the territory. Italian has absorbed influences from many of these dialects, as also from other languages due to dominion of portions of the country by other nations, such as Spain and France. 
  In this work we present Vulgaris, a project \draftMT{aimed at studying} a corpus of Italian textual resources from authors of different regions, ranging in a time period between 1200 and 1600. 
  Each composition is associated to its author, and authors are also grouped in families, i.e. sharing similar stylistic/chronological characteristics. 
  Hence, the dataset is not only a valuable resource for studying the diachronic evolution of Italian and the differences between its dialects, but it is also useful to investigate stylistic aspects between single authors.
  We provide a detailed statistical analysis of the data, and a corpus-driven study in dialectology and diachronic varieties.
\end{abstract}

\section{Introduction}\label{intro}
\blfootnote{Accepted at COLING Workshop on NLP For Similar Languages, Varieties And Dialects (VarDial) 2020, DOI: TBA}
% \blfootnote{
% \hspace{-0.65cm}
% This work is licensed under a Creative Commons Attribution 4.0 International Licence. Licence
% details: \url{http://creativecommons.org/licenses/by/4.0/}.}
% general pippone
Understanding the evolution of a language is a challenging problem. When a language originated? What are the influences coming from dialects and other languages? These are crucial questions that the study of language evolution aims to face. 

Natural Language Processing techniques are powerful tools that can support researchers in the analysis of dialects and diachronic language varieties \cite{zampieri2020similar,ciobanu2020automatic}. 
There exists several lines of research that approach the problem of defining distances between languages or varieties. Linguistic phylogenetics \cite{borin2013and} aim at determining a rooted tree to describe the evolution of a group of languages or varieties. Trees are built based on the so called \textit{lexicostatistics} technique, that takes into account words with common origin to determine a taxonomic organization of the languages.
Language distance approaches instead rely on the distributional hypothesis of words and require cross-lingual corpora. Similarity is based on word co-occurrences \cite{asgari2016comparing,liu2013language}, or using perplexity-based methods \cite{basile2016diachronic,gamallo2017perplexity,campos2018measuring,campos2020measuring}. Perplexity is estimated from Language Models, typically n-grams LMs of characters, trained on one corpus and evaluated on another variety.

Differently from previous work, we consider Neural Language Models (NLMs) \cite{bengio2003neural,mikolov2010recurrent}, that are more robust 
estimators well known for their generalization capabilities and currently the state-of-the-art approaches in Language Modeling tasks. There is a vast literature on NLMs. Many works also address the problem of character language modeling \cite{exploringlimitslm,hierachicalcharlm} or character-aware LMs \cite{marra2018unsupervised,charaware}.   

% our work
In this work we focus on Italian, a Romance language derived from Vulgar Latin. The uniquely fragmented political situation that occurred in Italy during the middle age makes Italian an extremely variegate and complex case of study, rich of dialects that are still spoken nowadays.
%We created \textit{Vulgaris}, a corpus of medieval text collections, 
We consider a corpus of medieval text collections, with the purpose of easing the research activity on diachronic varieties. Moreover, the dataset can be also a valid resource to study the problem Text Generation in low-data and variegate styles conditions. Similar corpora have been already collected for other languages. 
Colonia \cite{zampieri2013colonia}, is a  Portuguese diachronic dataset of about 5 milion tokens grouped by century in five sub-corpora.
In \cite{campos2020measuring} they gathered three corpora for English, Spanish and Portuguese. 

% contributions 
In summary, the contributions of this paper are: (1) we present a project, \textit{Vulgaris}, that studied a text corpus consisting of vulgar Italian language literary resources, organized in such a way to ease language research, (2) studying the historical and geographical background, the statistical properties of the collected data and its composition  and (3) deepening our analysis through  a corpus-driven study in dialectology and diachronic varieties exploiting perplexity-based distances. In particular, we introduce Neural Language Models to estimate the perplexity and provide a new indicator, named as Perplexity-based Language Ratio (PLR), to analyse the historical evolution process of the varieties.

% paper structure
The rest of the paper is organized as follows. In Section \ref{sec:hist}, we describe in detail \textit{Vulgaris}, its composition and we report several statistics on it. Then, in Section \ref{sec:exps}, we introduce perplexity-based metrics combined with neural language models that are then used to carry out experiments on the diachronic varieties within \textit{Vulgaris}.

\section{Vulgaris}\label{sec:dataset}
\label{sec:hist}
\draftMT{The main goal of project \textit{Vulgaris}\footnote{\draftAZ{The project is available at \url{https://sailab.diism.unisi.it/vulgaris/}.}} is the analysis of the diachronic evolution and variance of the vulgar italian language. In order to do so,   }
we collected an heterogeneous literary text corpus, comprehensive of poetry, prose, epistles and correspondence by the most important Italian authors ranging from the dawn of the vulgar language to the Reinassance Age. 
\draftAZ{Henceforth, for compactness, we refer to such data as \textit{Vulgaris}.}
The dataset represents a fundamental timeframe for the Italian language, including the first steps and diachronic evolutions departing from the Latin language. 
Moreover, through \textit{Vulgaris} it is possible to gain evidence of the early language fragmentation deriving from the complex historical  geo-political context of the Middle Age. 
% Indeed, 
% collezionato dataset eteronegeo, arco temporale ampio , evoluzione volgare temporale e territoriale.  frammentazione territoriale  ee politica sfocia i frammentazione linguistica, analisi molto ricca da questo punto di vista.
\subsection{Hystorical background and families}
The earliest years of the 13th century were characterized by a novel and complex civilisation. The rise of medieval Communes, associations among citizens of towns belonging to the same social class, influenced the rise of a novel school of secular thought increasingly unhindered by the religious influences. For these reasons, along with the establishment of the first universities, beside latin literature the vulgar Italian language started to appear in various literary works. The heterogeneous political and geographical context led to a linguistic fragmentation, characterized by various contact points.
The first literary evidence of vulgar poetic, which we denote as belonging to the \textbf{Archaic text} family,  is a collection of verses still connected with religious and moral themes, written in regions of the central Italy, in particular Umbria and Tuscany.  Amongst the main authors, we mention \textit{Francesco d'Assisi}. Inspired by this works, in the middle of the century (about 1250)  some vulgar authors (e.g. \textit{Jacopone da Todi} et al.), in the same geographical zone, composed several \textbf{Laude}, enriching the religious and mystical poetry theme. 

The \textbf{Northern Didactic poetry} family, flourished in the same years, was influenced by these religious and moral guidelines. We point out \textit{Bonvesin da la Riva} from Milan and \textit{
Giacomino da Verona} among the representatives, having the goal to instruct readers about morality, philosophy and doctrine.

The prosperous and thriving Imperial court of Federico II fostered the birth of a \textbf{Sicilian School} (1230-1250), where the figure of an angelic woman and the stereotype of love play a central role. This group laid the foundations of modern poetry, introducing a specific metric and organization in \textit{Stanzas}, creating Sonnets and unifying the language lexicon and structure.
Among the authors we highlight \textit{Giacomo da Lentini} and \textit{Pier della Vigna}.

With the death of Federico II, the cultural axis moved to Tuscany, thanks to the proliferation of Communes. 
Differently from the Sicilian School we cannot talk of a unique literary school. Indeed, in several important cities, such as Pisa, Lucca, Arezzo, Siena, other than Florence, which became only afterwards the most important cultural center, emerged themes inspired by the Sicilian similar ones. 

The \textbf{Northern/Tuscan Courtly poetry} arises from poets belonging to the Sicilian school who moved after the decadence of the Svevian Empire, influencing the themes and style of local authors (\textit{Guittone D'Arezzo}, \textit{Bonagiunta Orbicciani}, \textit{Compiuta Donzella}). In the meanwhile, \textbf{Central Italy Didactic poetry} (\textit{Brunetto Latini}) and \textbf{Realistic Tuscan poetry} (\textit{Cecco Angiolieri, Folgòre da San Gimignano, Cenne de la Chitarra}) emerged, differentiated by the themes, goal of the poetry, and style. Departing from the literature inspired by court life, a more popular and playful genre, the \textbf{Folk and Giullaresca Poetry}, was mainly due to jesters such as \textit{ Ruggieri Apuliese}.

Finally, thanks to the influence of Sicilian School and Tuscan poetry, the \textbf{Stilnovisti} family (\textit{Guido Guinizzelli}, \textit{ Dante Alighieri}, \textit{Guido Cavalcanti}, \textit{Lapo Gianni}, \textit{Gianni Alfani}, \textit{ Dino Frescobaldi}, \textit{Cino da Pistoia}) and some authors close to them (\textbf{Similar to Stilnovisti} - \textit{Lippo Pasci de Bardi}) evolved and refined the  poetry of their predecessors. Metaphors, a noble symbolism and introspection characterize this movement, which was  born in Bologna and developed in Florence reaching its climax.

\textbf{Boccaccio} and \textbf{Petrarca} compose, together with Dante, the three \textit{Crowns} of Italian literature. Their poetry and prose are inspired by \textit{Dolce Stil Novo}, with an evolution toward a more  wordily thematic, rather than spiritual. Their linguistic style is the offspring of an evolved society. 

The works developed by these families highly influenced following authors. In particular, \textbf{Ariosto} and \textbf{Tasso}, at the beginning of the 16th century,  were deeply inspired by Petrarca and the Stilnovisti, respectively. However, their different temporal context was reflected in their literary works. 

\begin{figure}[h]
\centering
      \includegraphics[ width=0.3\textwidth, trim={5cm 3cm 5cm 3cm},clip]{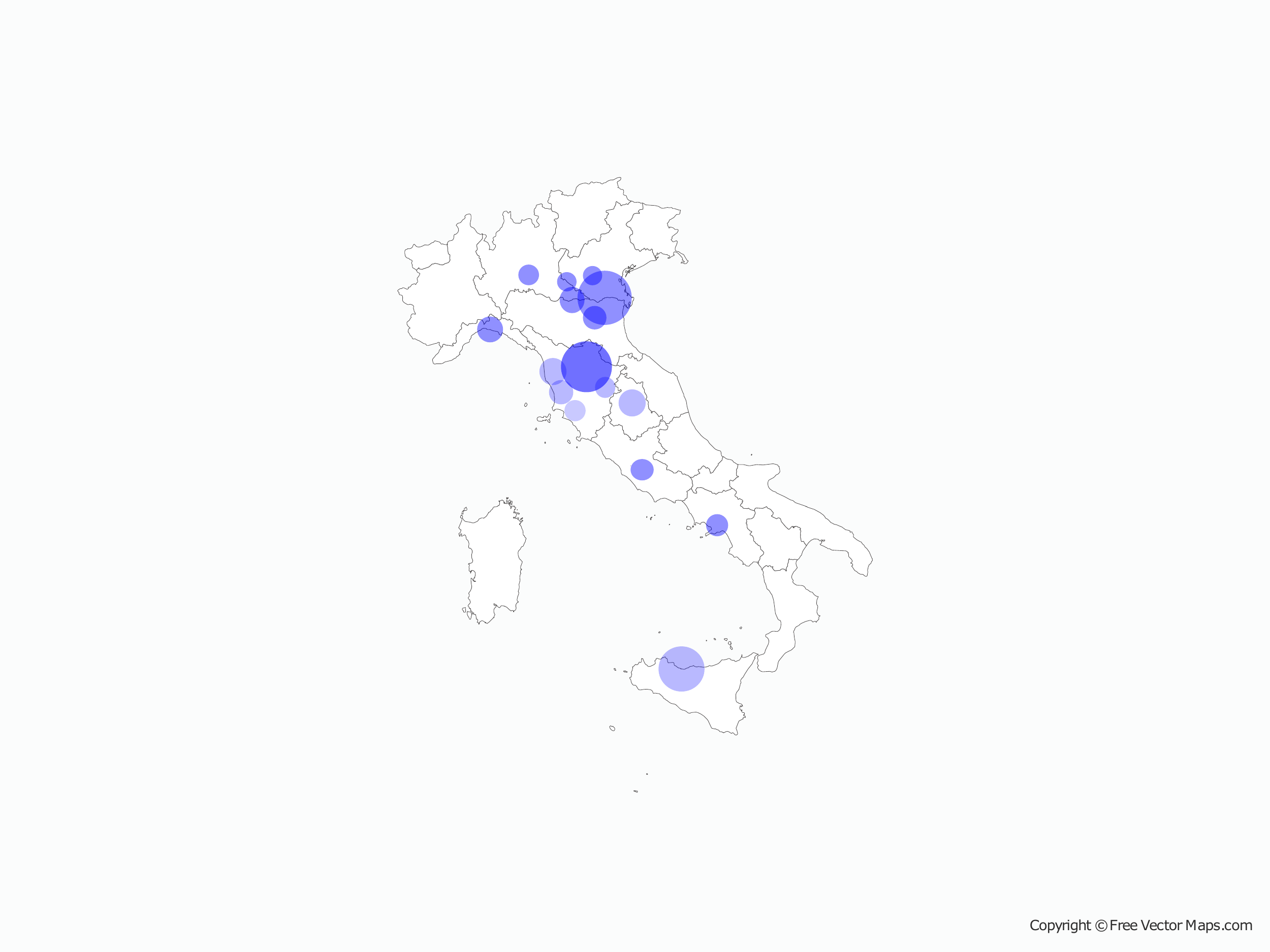}
      \includegraphics[width=0.69\textwidth]{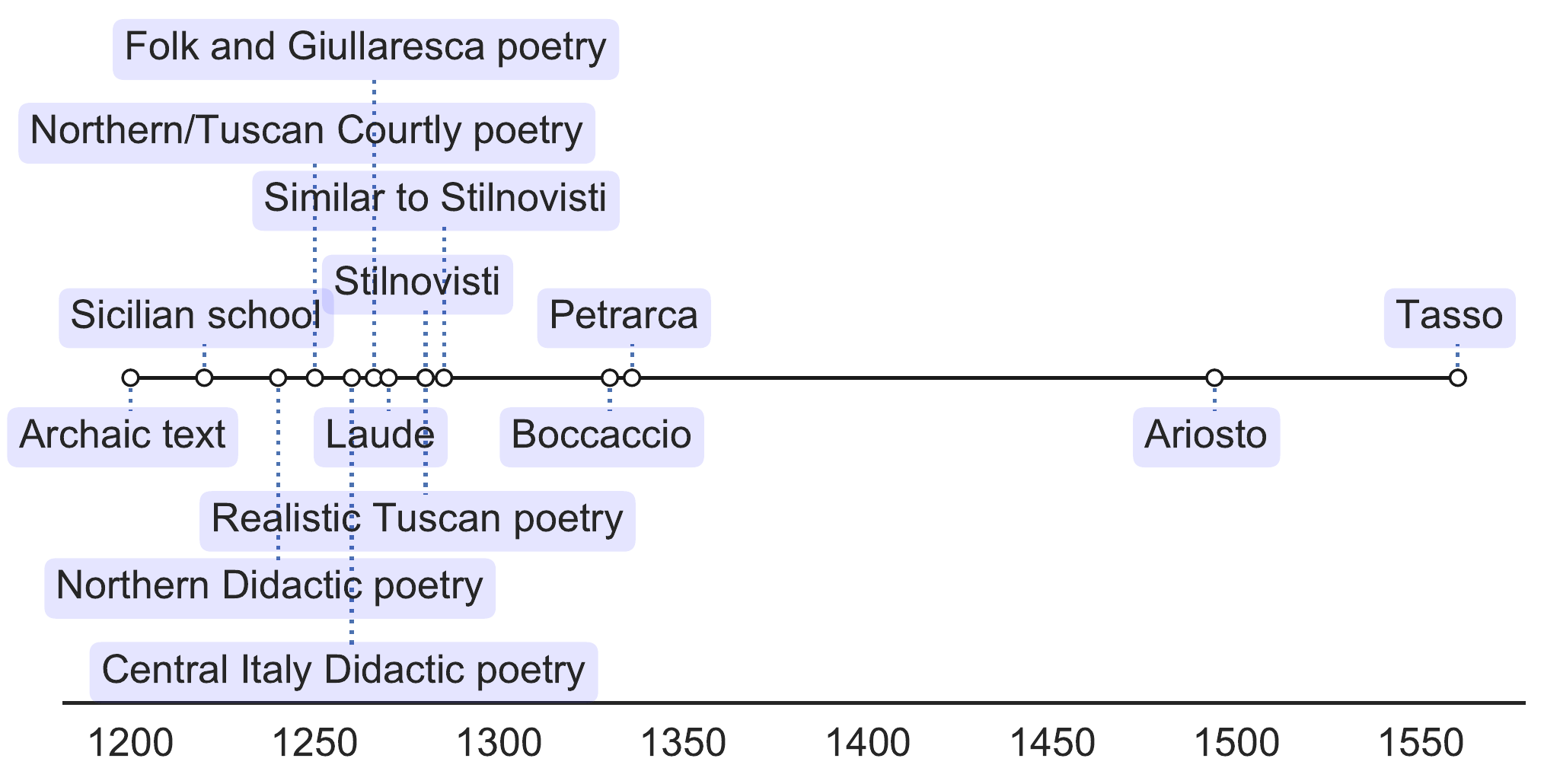}
 \caption{On the left side, a geographical map of the Italian peninsula with the biggest cultural centers marked  (the map is attributed to \url{https://freevectormaps.com/italy/IT-EPS-01-0004}). On the right, a timeline representing the temporal sequence of the different families we described in the main text.}
\label{fig:map}
\end{figure}

Therefore, the vulgar Italian language, starting from the beginning of the 13th century, became more and more popular amongst various authors, evolving during the following years in several families, which we summarize in Fig.\ref{fig:map}. The highly fragmented geo-political context gave rise to different schools, groups, communities (depicted in Fig. \ref{fig:map}) and hence many language varieties, dialects, that even nowadays are noticeable.  
%Hence, by providing the \draftMT{Vulgaris project} text corpora included in \textit{Vulgaris},

\draftMT{Through the Vulgaris project, we aim to provide a rich resource to analyze the diachronic evolution of the early Italian language, in particular in  poetry, prose and correspondence texts.}

% \subsection{Dataset structure}
% \textcolor{red}{TENERE SE SI DECIDE CHE COPYRIGHT OK; ALTRIMENTI SOLO DIRE CHE SI è PARSATO IL WEB?}
%  The available text are released under a Creative Commons license, CC BY-NC-ND 2.0 IT, which guarantee the possibility to freely reuse the resources requiring attribution and without adaptation. 
% %We extracted the texts solely from this resource, hence we respect the copyright requirements.
 
% We crawled the XML/TEI encoded texts through an ad-hoc custom parser, built upon Python and XPath library. 

\subsection{Dataset structure}
The \draftMT{examined} corpus contains texts retrieved from Biblioteca Italiana\footnote{\url{http://bibliotecaitaliana.it/}}, a digital library project collecting the most significant texts of the Italian literature, ranging from the Middle Age to the 20th century. 
\draftAZ{The code to retrieve and analyse the data can be found in \url{https://github.com/sailab-code/vulgaris}.}
% IN the folowing, we refer to Vulgaris as the corpus we collected...
\textit{Vulgaris} provides the following filtered type of information, extracted from  the parsed data: \textbf{author},  \textbf{title},  \textbf{collection}, \textbf{family},  \textbf{type}, \textbf{text}.
In details, the corpus is composed by \textbf{text} produced by $104$ \textbf{authors} belonging to the $14$ \textbf{families}  described in Section \ref{sec:hist}. The corpus contains $177$ \textit{collections}, consisting of groups of poetry, single poems, personal epistles. Each item of the dataset is a single composition, for instance a poetry, a chapter or a letter. Moreover, we split the resources by the \textbf{style} attribute into \textit{poetry} and \textit{prose}, with the latter containing the both the prose and correspondence documents.

The structure of a poetic composition represents an important information in tasks such as Poem Generation \cite{lau2018deep,zugarini2019neural,zhang2014chinese}. The verse organization of the poetry is encoded by tags denoting each line break \texttt{<EOL>} (end of a verse), as well as the end of each stanza \texttt{<EOS>}.
In the case of prose, only the organization in paragraphs is represented by the tag \texttt{<EOS>}.

In Table \ref{tab:stats} we report some statistics on the families (first column, ordered by date), including their most representative authors (second column), the total amount of collected texts, divided into poetry and prose (third, fourth and fifth column, respectively). Whilst the older families are underrepresented, families belonging to a later period are mostly characterized by a larger amount of texts. This fact is a good indicator of the diffusion that the Italian language  has undergone during this timeline. 
% is a good proxy to better point out how the evolution and diffusion of  fostered the flourishing of prolific authors.

\begin{table}[]
\small
    \centering
    \begin{tabular}{c|c|c|c|c}
        \toprule    
        \textit{Family}  & \textit{Authors} &  \textit{\#Texts} &  \textit{\#Poetry}  &  \textit{\#Prose} \\
        \midrule
         \textit{Archaic text} & \shortstack{Francesco d'Assisi, Ritmo Laurenziano, \\ Ritmo Cassinese }   &    5 &  5 & - \\
         \midrule
         \textit{Sicilian School} & \shortstack{Giacomo da Lentini, Guido delle Colonne,\\ Pier della Vigna, Pronotaro da Messina}    &   46 &  46 & - \\
         \midrule
         \textit{\shortstack{Northern Didactic \\ poetry}} & \shortstack{Girardo Patecchio Da Cremona, Bonvesin Da La Riva,\\ Giacomino Da Verona, Anonimo Genovese}    &   29 &  29 & - \\
        \midrule
        \textit{\shortstack{Northern/Tuscan \\Courtly poetry }} & \shortstack{Guittone D'Arezzo, Bonagiunta Orbicciani,\\ Chiaro Davanzati, Monte Andrea Da Firenze}    &   101 &  101 & - \\
        \midrule
         \textit{\shortstack{Central Italy \\Didactic poetry  }} & \shortstack{Brunetto Latini,Garzo, Detto Del Gatto Lupesco\\ Dal Bestiario Moralizzato Di Gubbio}    &   8 &  8 & - \\
        \midrule
        \textit{\shortstack{Folk and Giullaresca \\ poetry    }} & \shortstack{Ruggieri Apugliese, Castra Fiorentino,\\ Matazone Da Caligano, Rime Dei Memoriali Bolognesi}    &  23 &  23 & - \\
        \midrule
        \textit{\shortstack{Laude}} & \shortstack{Jacopone Da Todi, Laude Cortonesi,\\ Lauda Dei Servi Della Vergine}    &  41 &  41 & - \\
        \midrule
        \textit{\shortstack{Stilnovisti}} & \shortstack{Guido Guinizzelli, Guido Cavalcanti, \\Cino da Pistoia, Dante Alighieri, Lapo Gianni}    &  769  &  704 & 65 \\
        \midrule
        \textit{\shortstack{Realistic Tuscan \\poetry}} & \shortstack{Rustico Filippi, Cecco Angiolieri,\\ Folgore da San Gimignano, Cenne de la Chitarra}    &  69  &  69 & - \\
        \midrule
        \textit{\shortstack{Similar to \\Stilnovisti}} & \shortstack{Dante's Friend, \\Lippo Pasci de' Bardi}    &  709  &  70 & - \\
        \midrule
        \textit{\shortstack{Boccaccio}} & \shortstack{-}    &  1058  &  296 & 762 \\
        \midrule
        \textit{\shortstack{Petrarca}} & \shortstack{-}    &  872  &  747 & 125 \\
         \midrule
        \textit{\shortstack{Ariosto}} & \shortstack{-}    &  363  &  144 & 219\\
        \midrule
        \textit{\shortstack{Tasso}} & \shortstack{-}    &  3366  &  1604 & 1762\\
        \bottomrule
         \textbf{\shortstack{Total}} & \shortstack{-}    &  6820  &  3887 & 2933\\
    \end{tabular}
        \caption{Analysis of the composition of the dataset. We report the families, their most representative authors, total number of provided texts and their distribution in poetry and prose.}
    
    \label{tab:stats}
\end{table}

% statistiche globali, numero parole, numero componimenti. componimenti per autore, per famiglia. paolre per autore/ famiglia. 

% plot  +descrizione figura

% tabelle con familgia ed autori (principali). Autori principali ed Opere

% Caratteristiche vocabolario (differenze sicilaino da dolce stil noivo, testi arcaici) 

% opere principlai e numero di parole. 
% lunghezza massima di un testo. 

% paisa https://pdfs.semanticscholar.org/8dfd/c63897da53c737a6f357e428534efea7a50a.pdfhttps://www.overleaf.com/project/5f004f89a7fe680001429846
\draftMT{The corpus investigated in  \textit{Vulgaris} is extremely heterogeneous and}  composed by 4 million word occurrences, whose texts have been written by authors from a wide range of geographical regions and time periods, as shown in Figure \ref{fig:map}. 
 In Table \ref{tab:total_word} we summarize some statistics on the total amount of word occurrences, the number of unique words and the average occurrences per word for each text \textbf{type}. The total number of words in poetry and prose is almost balanced, whereas their composition is remarkably different. Indeed, poetry has a richer lexicon than prose, containing almost twice unique words.
 
\begin{figure*}[!ht]
\centering
       \includegraphics[ height=5cm]{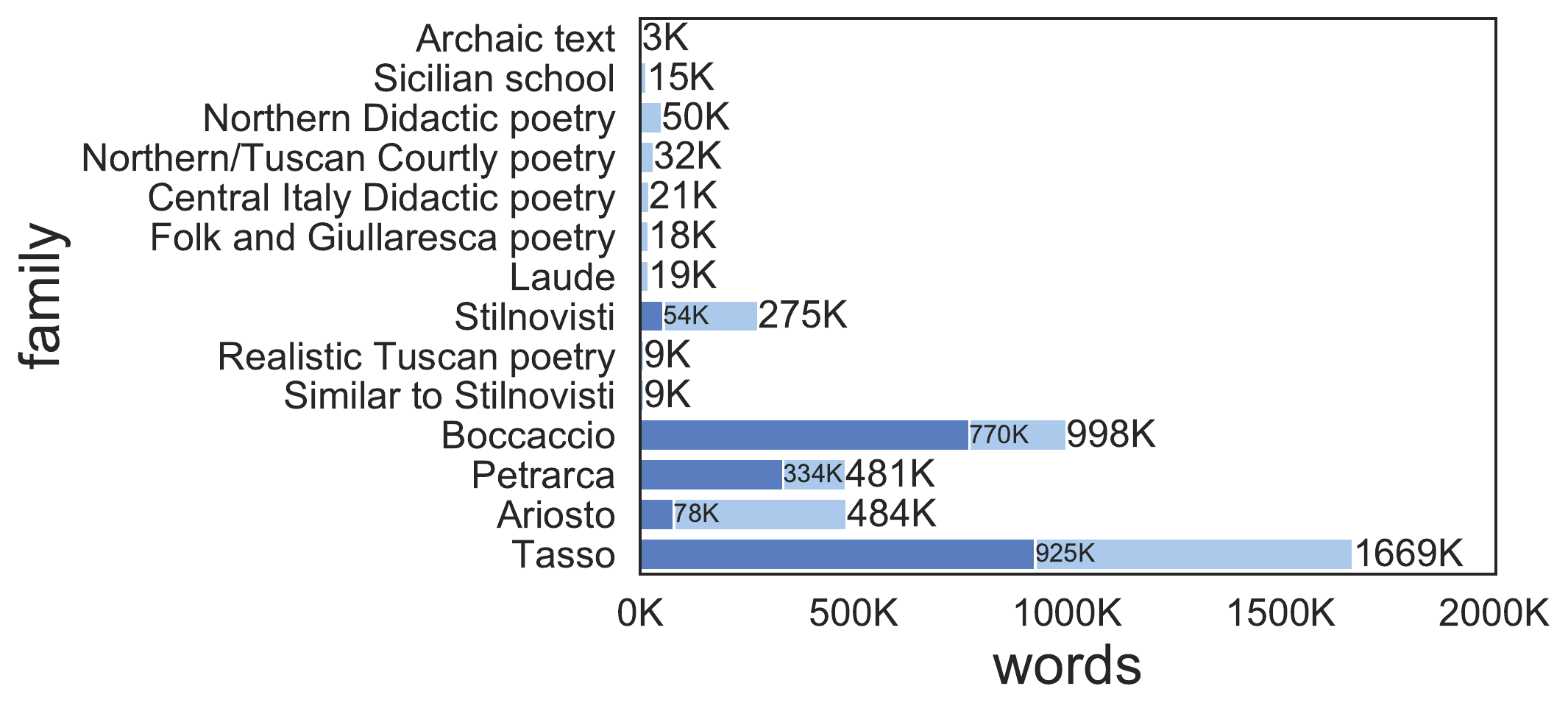}
 \caption{The figure reports the total amount of word occurrences for each family (both in poetry and prose), at the right of each family bin. The darker blue bar denotes the portion of word occurrences in prose texts only.}
\label{fig:words}
\end{figure*}

 To depict the contribution of each family to the dataset, Figure \ref{fig:words} reports the total number of word occurrences for each family and the poetry/prose proportion.
 Once more, these statistics confirm the increasing spread of the Italian language. We can also notice how vulgar spread. Initially, it was mainly used in poetry and only later vulgar prosaic forms appeared. Only 5 out of 14 families contain prose, and, as we can see from the timeline in Fig. \ref{fig:map}, they correspond to the latest families.  
 
\begin{table}[!ht]
%\small
    \centering
    \begin{tabular}{c|c|c|c}
        \toprule    
        & \textit{Global}  & \textit{Poetry} &  \textit{Prose} \\
        \midrule
        \# word occurrences & 4090166 & 1925838 & 2164328\\
        \# unique words & 180450 & 136195 & 69135\\ \hline
        Avg occurrences per word & 22.67 & 14.14 & 31.31
    \end{tabular}
        \caption{Statistics on words for each text category.}
    \label{tab:total_word}
\end{table}

\begin{figure*}[!ht]
\centering
    \includegraphics[ height=4.2cm]{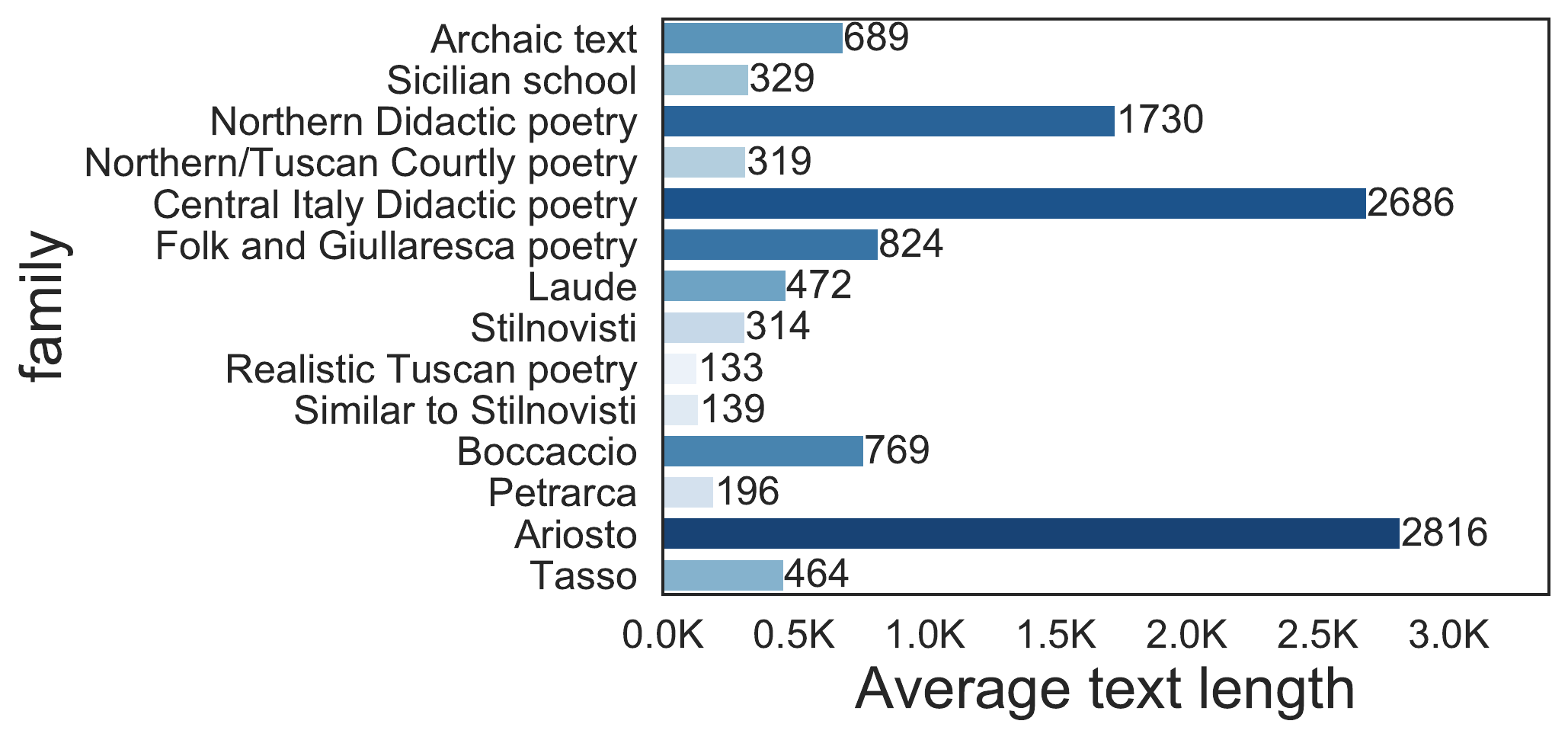}
    \includegraphics[height=4.2cm]{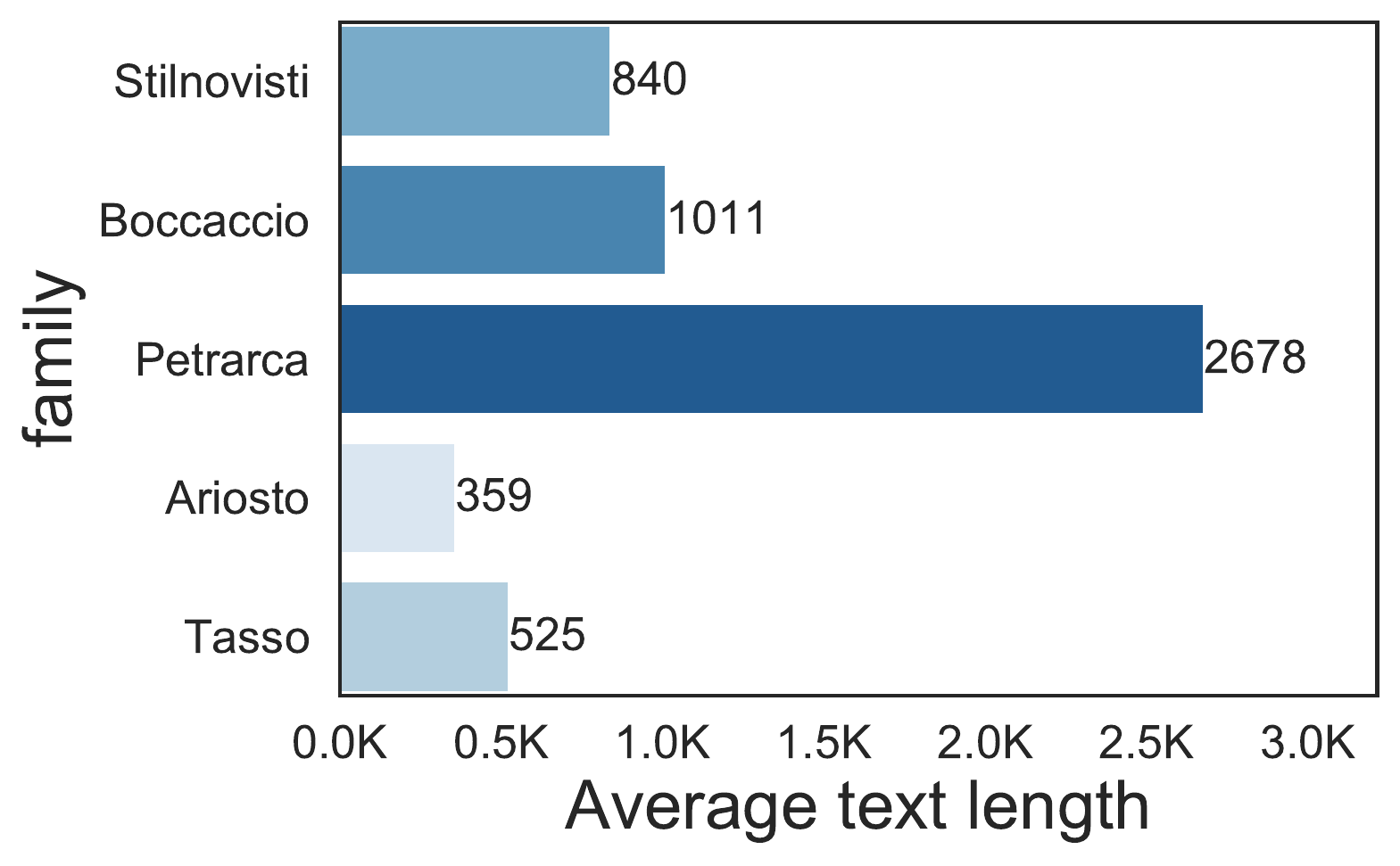} 
    \includegraphics[ height=4.2cm]{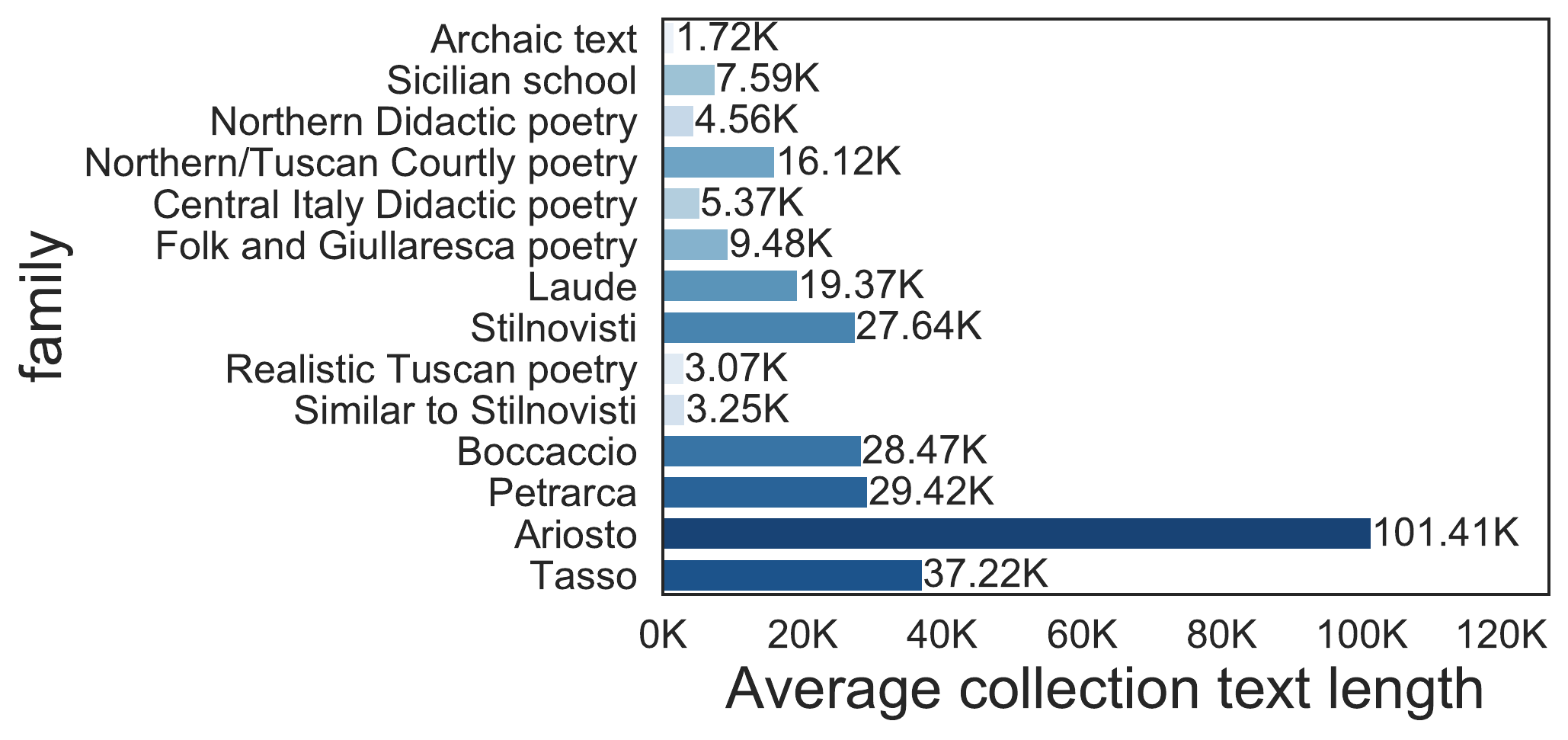}
    \includegraphics[height=4.20cm]{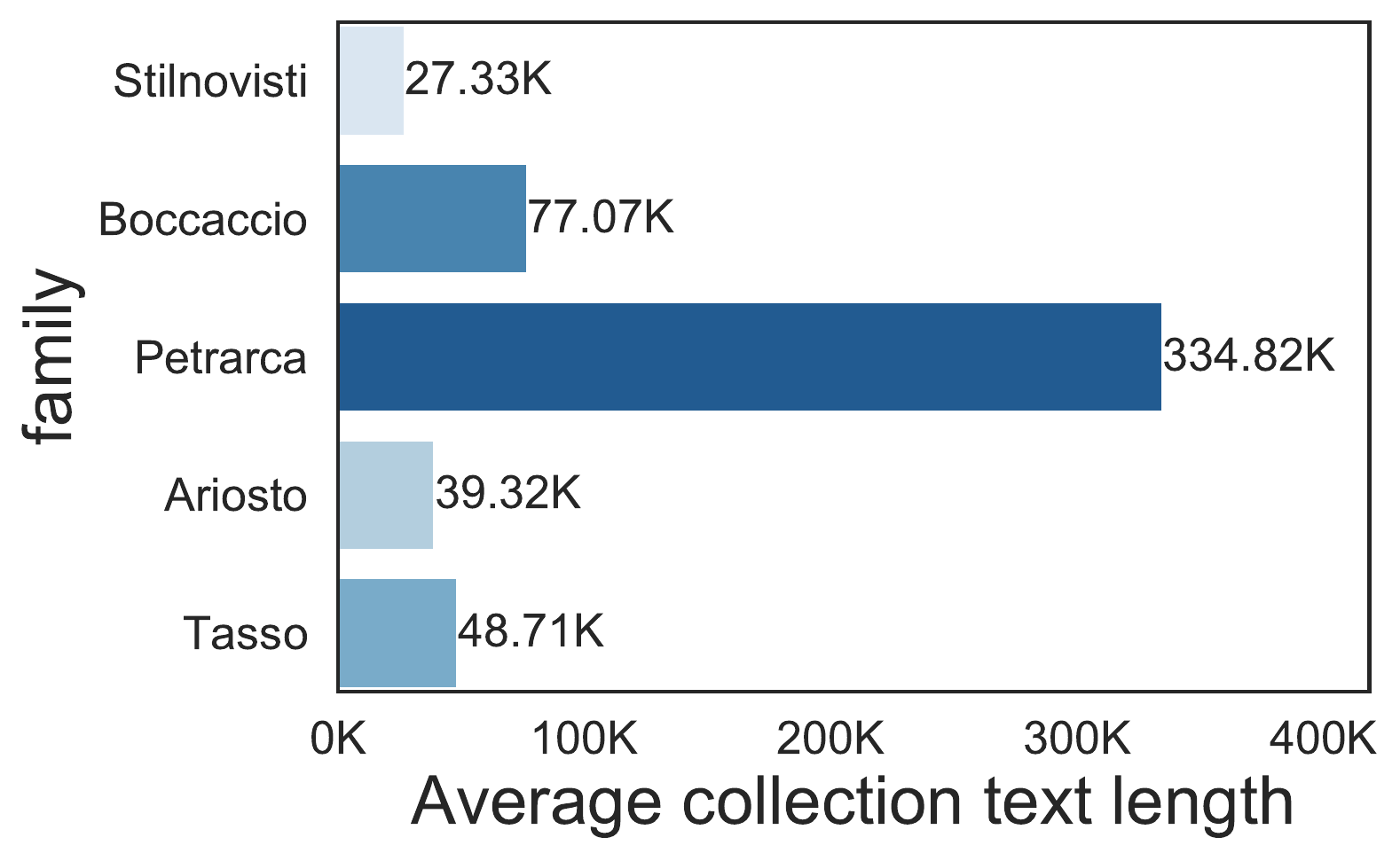} 
 \caption{In the top row, the average text length of poetry (left) and prose (right) texts. On the bottom, average number of words contained in each collection in poetry (left) and prose (right). }
\label{fig:avg_txt}
\end{figure*}

Finally, in the top row of Fig.~\ref{fig:avg_txt}, we report the average distribution of the text length, in both the styles (i.e, \textit{poetry} on the left and \textit{prose} on the right) among all the families.  The bottom row of Fig.~\ref{fig:avg_txt} shows the  average number of words contained in each collection, hence texts having similar characteristics or theme.

% \begin{figure*}
%     \centering
% %  \includegraphics[width=0.4\textwidth]{poems4family_prose.pdf}
%         \caption{Caption}
%     \label{fig:my_label}
% \end{figure*}

\section{Analysis of Language Varieties in Vulgaris}\label{sec:exps}
 \textit{Vulgaris} texts span over a time period of about four centuries. The diachronic varieties within the dataset are measured in terms of perplexity-based distances, taking into account the different centuries as a reference for the comparison.

\subsection{Perplexity-based Language Distance}\label{subsec:pld}
The quality of a Language Model (LM) is assessed in terms of perplexity. A good estimation by a language model for a given corpus will yield low perplexity values. LMs and perplexity have been already exploited to provide a distance between language corpora \cite{gamallo2017perplexity}, and this approach has been effectively applied for language discrimination and the analysis of historical varieties \cite{campos2018measuring,campos2020measuring}. 

Let us consider two language corpora, namely $L1$ and $L2$, and let $\texttt{LM}_{L1}, \texttt{LM}_{L2}$ be two language models trained on $L1$ and $L2$, respectively. We can argue that the more the corpora are related to each other, the more accurate is the estimate provided by the LM trained on one language when evaluated on the other. By denoting the two measures of perplexity as $pp_{L1 \rightarrow L2}(L2, \texttt{LM}_{L1})$ and of $pp_{L2 \rightarrow L1}(L1, \texttt{LM}_{L2})$, the Perplexity-based Language Distance (PLD) is defined in \cite{gamallo2017perplexity} as the average of these two values:
\begin{equation}
    PLD(L1, L2) = \frac{pp_{L1 \rightarrow L2}(L2, \texttt{LM}_{L1}) + pp_{L2 \rightarrow L1}(L1, \texttt{LM}_{L2})}{2}.
\end{equation}
This metric copes with the fact that $pp_{L1 \rightarrow L2}(L2, \texttt{LM}_{L1})$ and $pp_{L2 \rightarrow L1}(L1, \texttt{LM}_{L2})$ are not symmetric, mostly because LMs are trained and tested on different data distributions. However, the asymmetry in these values can be a good indicator of the language evolution on diachronic/dialect varieties, since it can enlighten either a language compression/simplification or a language expansion over time.  Indeed, in the process of language unification, words are reduced, and dialectal expressions are suppressed, thus reducing the overall richness of the language. Hence, we consider also the following Perplexity-based Language Ratio (PLR): 
\begin{equation}
    PLR(L1, L2) = \frac{pp_{L1 \rightarrow L2}(L2, \texttt{LM}_{L1})}{pp_{L2 \rightarrow L1}(L1, \texttt{LM}_{L2})}.
\end{equation}
PLR values greater than 1 indicate that $L1$ is likely to be a more various language than $L2$, whereas values less than 1 are likely to indicate $L2$ as the more complex language.

\subsection{Conditional Language Modeling}\label{subsec:nlm}
In this section we briefly describe the structure of the language models that have been exploited in the experimental evaluation. Let us consider a sequence of tokens $\Bx = (x_1, \ldots, x_{n})$ from a text corpus. The following description is general for any sequence of tokens $\Bx$ regardless their kind, e.g. words, characters or any token piece. 
The goal of the LM is to estimate the joint probability $p(\Bx)$, that is factorized with the product of conditional probabilities as follows,
\begin{equation}
    p(\Bx) = \prod_{i=1}^{m} p(x_i|x_{i-1}, \ldots, x_1),
    \label{eq:lm}
\end{equation}
A Neural Language Model (NLM) \cite{bengio2003neural} estimates the conditional probability $p(x_i|x_{i-1}, \ldots, x_1)$ in Equation~(\ref{eq:lm}) with a Neural Network.
% Machine Translation, Text Summarization, Text Continuation, Poem Generation, and in general any sequence-to-sequence problem in NLP can be formulated as in Eq. (\ref{eq:nlg}).
We extend Equation~(\ref{eq:lm}) by adding other features of the text to condition the NLM. In particular, we leverage the external meta information about author $a$, family $f$ and kind of composition $k$ (prose or poetry) available in the dataset. Hence Equation~(\ref{eq:lm}) becomes:
\begin{equation}
    p(\Bx) = \prod_{i=1}^{m} p(x_i|x_{i-1}, \ldots, x_1, a, f, k).\label{eq:conditional_nlg}
\end{equation}

We model the distribution in Equation \ref{eq:conditional_nlg} by means of a recurrent neural network. Each token from the vocabulary $V$ of size $|V|$ is associated to a latent embedding $\Be$ of dimension $d$. The set of the $|V|$ embeddings are collected in the $|V|\times d$ matrix $\mathbf{E}$. 
In particular, we consider an LSTM cell \cite{hochreiter1997long} to model the internal recurrent state of the network. At time $t$ the state $\Bh_t$ is updated as follows,
\begin{equation}
	\Bh_{t} =  \texttt{LSTM}(\Be_{t}, \Bh_{t-1}) \ ,
\end{equation}
The external features ($a,f,k$) are concatenated to $\Bh_t$ and then linearly projected into a $d$-dimensional vector $\Bs_t$:
\begin{eqnarray*}
    \Bc_t = [\Bh_t \circ \Ba \circ \Bf \circ \Bk],\\
    \Bs_t = W \cdot \Bc_t  + b,
\end{eqnarray*}
where $\circ$ is the concatenation operator, and $\Ba, \Bf, \Bk$ the embedding representations associated to author $a$, family $f$ and kind $k$, respectively. 
The probability distribution $\hat{\By}_t$ is the output of a softmax layer sharing the weights of the input embeddings to apply a back-projection of the contextual state $\Bs_t$ into the vocabulary space: 
\begin{eqnarray*}
\Bo_t = \mathbf{E}^T \cdot \Bs_t,\\
\hat{\By}_t = \texttt{softmax}(\Bo_t).
\end{eqnarray*}

\begin{figure*}
\centering
      \includegraphics[width=0.7\textwidth, trim={0cm 3cm 0cm 3cm}, clip]{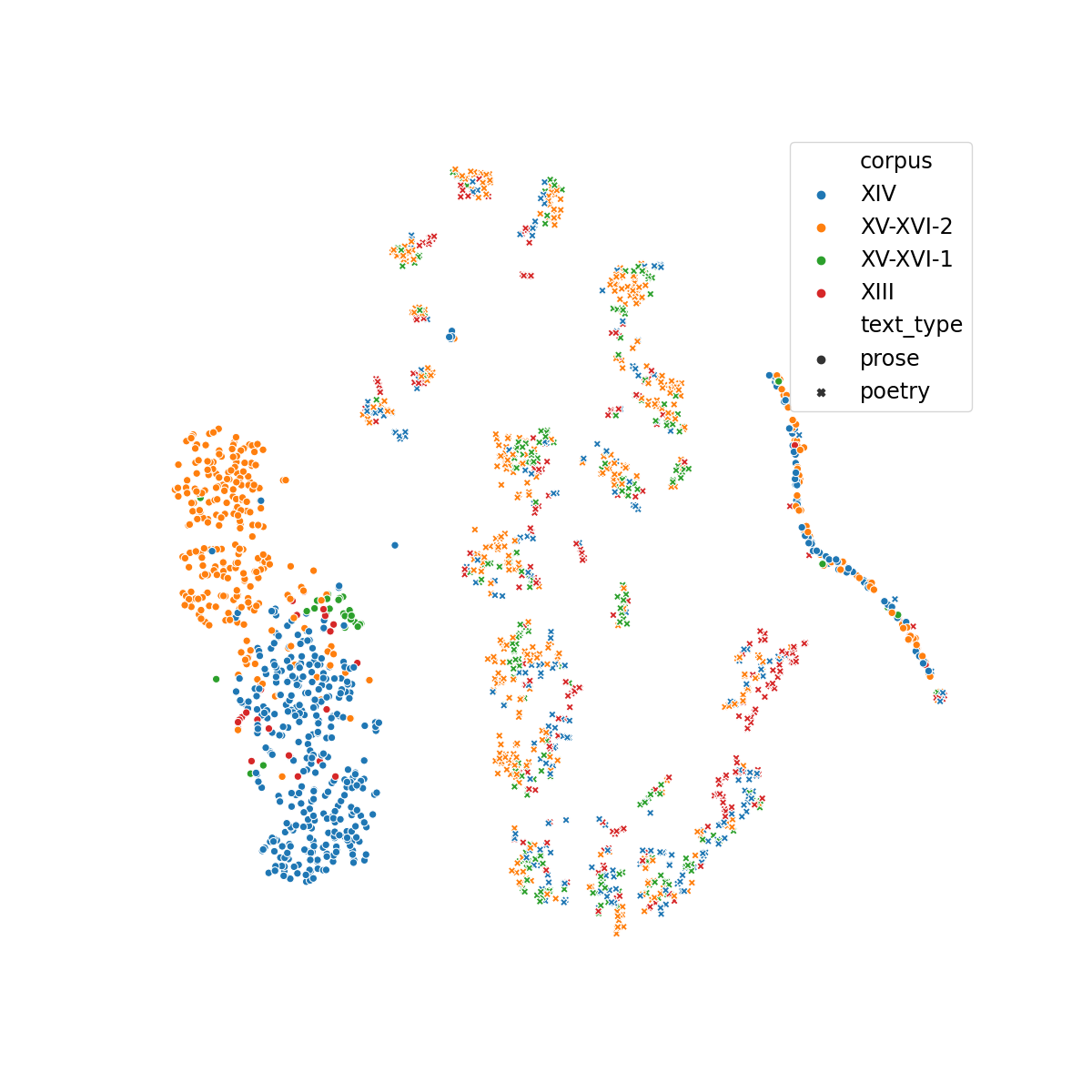}
 \caption{Two dimensional t-SNE representation of sentences' state $h_T$. Different colours indicate different groups, dots for poetry, crosses for prose.}
\label{fig:tsne}
\end{figure*}

The PLD and PLR are estimated exploiting this conditional neural language model where input tokens are characters. We chose NLMs over n-grams because of their notorious generalization capabilities. More robust LMs estimation will improve the quality of the PLD and PLR measures.
The same kind of architecture is used to build and visualize the sequence representations shown in Fig.~\ref{fig:tsne}, learnt from a word-based language model on the entire \textit{Vulgaris}.

\subsection{Diachronic Variety}
The $14$ families of \textit{Vulgaris} are arranged in four language corpora, based on their time periods, as shown in Fig.~\ref{fig:map}. The first group, referred to as XIII, includes all the families belonging to the 13th century. In this century there are $10$ out of $14$ families of the dataset, making this language variety the most heterogeneous one, including many authors from different areas of the Italian territory. In the second one (XIV), we consider \textit{Petrarca} and $\textit{Boccaccio}$ families/authors, whereas \textit{Ariosto} and \textit{Tasso} constitute the third and forth corpora, respectively, XV-XVI-1 and XV-XVI-2.
Clearly the boundaries are not neat, since the activity of some authors may span across two centuries.  
From Table \ref{tab:groups_stats} we can see that the diachronic corpora are unbalanced. Despite the high number of families and authors, the XIII corpus is the less represented one, followed by XV-XVI-1 that is slightly larger. They both are small compared to XIV and XV-XVI-2. However, XIII is also the dataset with lowest average number of occurrences per word, indicating a high variance of the collection caused by the rich variety of styles and authors.
\begin{table}[!ht]
\centering
\begin{tabular}{l|cccc}
\toprule
& XIII & XIV & XV-XVI-1 & XV-XVI-2   \\ 
\midrule
\# words & 455583 & 1480379 & 484276  & 1669928\\
 dataset proportion (\%) & 11.14 & 36.19  & 11.84 & 40.83\\ 
 \midrule
 \# unique words & 57343 & 73530  & 42594 & 72369\\
 Avg occurrences per word & 7.94 & 20.13  & 11.37 & 23.08
\end{tabular}\caption{Number of words and proportions of the four diachronic groups. }\label{tab:groups_stats}
\end{table}

As a first qualitative analysis, we trained a word-based conditional NLM on the
entire corpus, using a vocabulary of $50000$ words. The final cell state $h_T$ of a text sequence $x=(x_1, \cdots, x_T)$ is projected into a 2-dimensional representation using t-SNE \cite{maaten2008visualizing}. Fig.~\ref{fig:tsne} visualizes the 2-d representation of $2000$ examples, colored accordingly to the corpus they belong to, and styled differently in case of prose or poetry works. Prose and poetry are easily discriminated by the NLM. The corpus origin is also captured by the NLM, although not completely, suggesting that the diachronic varieties share a similar structure. 

Then, both the PLD and the PLR described in Subsection \ref{subsec:nlm} are computed for each pair of corpora. For the character LMs, we consider input character sequences with a maximum length of $50$. The state $h_t$ has size $256$, with $(\Ba,\Bf, \Bk)$ of size $16$, $16$ and $32$, respectively. Special tokens delimiting end of sentence, end of verse and white space are included in the vocabulary of characters. 
For each $L_i \rightarrow L_j$, the network is trained on $90\%$ of the $L_i$ corpus, whereas the remaining $10\%$ is used for early stopping, and it is finally evaluated on the whole $L_j$. 
\begin{table}[!ht]
\centering
\begin{tabular}{l|cccc}
\toprule
 & XIII & XIV & XV-XVI-1 & XV-XVI-2 \\ 
 \midrule
 XIII & 3.90 & 5.38 & 5.99 & 6.08  \\
 XIV & 5.38  & 3.52 & 4.76  & 4.65 \\
 XV-XVI-1 & 5.99 & 4.76 & 3.30 & 4.47 \\
 XV-XVI-2 & 6.08 & 4.65 & 4.47 & 3.28\\ 
\end{tabular}\caption{PLD among pairs of diachronic language varieties.}\label{tab:pld_groups}
\end{table}

Results are shown in tables \ref{tab:pld_groups} and \ref{tab:ppl_ratio_groups}. PLD is lower in diachronic varieties closer in time, as expected.
Interestingly enough, PLR highlights a strong asymmetric behaviour on perplexity pairs involving the set XIII. Indeed, while training a language model on a heterogeneous corpus, as it is XIII, makes the LM well performing when testing on simpler varieties, a language model trained on a poorer corpus underperforms when evaluating it on a richer corpus, as XIII. 

\begin{table}[!ht]
\centering
\begin{tabular}{l|cccc}
\toprule
 & XIII & XIV & XV-XVI-1 & XV-XVI-2 \\ 
 \midrule
 XIII & 1.00 & 0.81 & 0.65 & 0.72  \\
 XIV & \textbf{1.23}  & 1.00 & 0.86  & 0.95 \\
 XV-XVI-1 & \textbf{1.53} & 1.16 & 1.00 & 1.14 \\
 XV-XVI-2 & \textbf{1.39} & 1.05 & 0.88 & 1.00\\ 
\end{tabular}\caption{PLR among pairs of diachronic language varieties.}\label{tab:ppl_ratio_groups}
\end{table}

\section{Conclusions}
In this paper we described {\em Vulgaris}, \draftAZ{a project that analyzes} a collection of literary texts covering the production of Italian authors mainly from the middle age. The dataset contains both poetry and prose, and each document is enriched by metadata that provide both information on the text characteristics and structure (the verse and stanza organization for poems and the paragraph splitting for prose). A preliminary analysis on the dataset by means of both simple statistics and perplexity--based measures gives some insights on the main feature of the collection that reflects the complexity and diachronic properties of Italian language in the early stages of its birth.

\bibliographystyle{coling}
\bibliography{coling2020}

\begin{thebibliography}{}

\bibitem[\protect\citename{Asgari and Mofrad}2016]{asgari2016comparing}
Ehsaneddin Asgari and Mohammad~RK Mofrad.
\newblock 2016.
\newblock Comparing fifty natural languages and twelve genetic languages using
  word embedding language divergence (weld) as a quantitative measure of
  language distance.
\newblock {\em arXiv preprint arXiv:1604.08561}.

\bibitem[\protect\citename{Basile \bgroup et al.\egroup
  }2016]{basile2016diachronic}
Pierpaolo Basile, Annalina Caputo, Roberta Luisi, and Giovanni Semeraro.
\newblock 2016.
\newblock Diachronic analysis of the italian language exploiting google ngram.
\newblock {\em CLiC it}, page~56.

\bibitem[\protect\citename{Bengio \bgroup et al.\egroup
  }2003]{bengio2003neural}
Yoshua Bengio, R{\'e}jean Ducharme, Pascal Vincent, and Christian Jauvin.
\newblock 2003.
\newblock A neural probabilistic language model.
\newblock {\em Journal of machine learning research}, 3(Feb):1137--1155.

\bibitem[\protect\citename{Borin}2013]{borin2013and}
Lars Borin.
\newblock 2013.
\newblock The why and how of measuring linguistic differences.
\newblock {\em Approaches to measuring linguistic differences, Berlin, Mouton
  de Gruyter}, pages 3--25.

\bibitem[\protect\citename{Campos \bgroup et al.\egroup
  }2018]{campos2018measuring}
Jos{\'e} Ramom~Pichel Campos, Pablo Gamallo, and I{\~n}aki Alegria.
\newblock 2018.
\newblock Measuring language distance among historical varieties using
  perplexity. application to european portuguese.
\newblock In {\em Proceedings of the Fifth Workshop on NLP for Similar
  Languages, Varieties and Dialects (VarDial 2018)}, pages 145--155.

\bibitem[\protect\citename{Campos \bgroup et al.\egroup
  }2020]{campos2020measuring}
Jos{\'e} Ramom~Pichel Campos, Pablo~Gamallo Otero, and I{\~n}aki~Alegria
  Loinaz.
\newblock 2020.
\newblock Measuring diachronic language distance using perplexity: Application
  to english, portuguese, and spanish.
\newblock {\em Natural Language Engineering}, 26(4):433--454.

\bibitem[\protect\citename{Ciobanu and Dinu}2020]{ciobanu2020automatic}
Alina~Maria Ciobanu and Liviu~P Dinu.
\newblock 2020.
\newblock Automatic identification and production of related words for
  historical linguistics.
\newblock {\em Computational Linguistics}, 45(4):667--704.

\bibitem[\protect\citename{Gamallo \bgroup et al.\egroup
  }2017]{gamallo2017perplexity}
Pablo Gamallo, Jos{\'e} Ramom~Pichel Campos, and Inaki Alegria.
\newblock 2017.
\newblock A perplexity-based method for similar languages discrimination.
\newblock In {\em Proceedings of the fourth workshop on NLP for similar
  languages, varieties and dialects (VarDial)}, pages 109--114.

\bibitem[\protect\citename{Hochreiter and Schmidhuber}1997]{hochreiter1997long}
Sepp Hochreiter and J{\"u}rgen Schmidhuber.
\newblock 1997.
\newblock Long short-term memory.
\newblock {\em Neural computation}, 9(8):1735--1780.

\bibitem[\protect\citename{Hwang and Sung}2017]{hierachicalcharlm}
Kyuyeon Hwang and Wonyong Sung.
\newblock 2017.
\newblock Character-level language modeling with hierarchical recurrent neural
  networks.
\newblock In {\em Acoustics, Speech and Signal Processing (ICASSP), 2017 IEEE
  International Conference on}, pages 5720--5724. IEEE.

\bibitem[\protect\citename{Jozefowicz \bgroup et al.\egroup
  }2016]{exploringlimitslm}
Rafal Jozefowicz, Oriol Vinyals, Mike Schuster, Noam Shazeer, and Yonghui Wu.
\newblock 2016.
\newblock Exploring the limits of language modeling.
\newblock {\em arXiv preprint arXiv:1602.02410}.

\bibitem[\protect\citename{Kim \bgroup et al.\egroup }2016]{charaware}
Yoon Kim, Yacine Jernite, David Sontag, and Alexander~M Rush.
\newblock 2016.
\newblock Character-aware neural language models.
\newblock In {\em AAAI}, pages 2741--2749.

\bibitem[\protect\citename{Lau \bgroup et al.\egroup }2018]{lau2018deep}
Jey~Han Lau, Trevor Cohn, Timothy Baldwin, Julian Brooke, and Adam Hammond.
\newblock 2018.
\newblock Deep-speare: A joint neural model of poetic language, meter and
  rhyme.
\newblock In {\em Proceedings of the 56th Annual Meeting of the Association for
  Computational Linguistics (Volume 1: Long Papers)}, pages 1948--1958.

\bibitem[\protect\citename{Liu and Cong}2013]{liu2013language}
HaiTao Liu and Jin Cong.
\newblock 2013.
\newblock Language clustering with word co-occurrence networks based on
  parallel texts.
\newblock {\em Chinese Science Bulletin}, 58(10):1139--1144.

\bibitem[\protect\citename{Maaten and Hinton}2008]{maaten2008visualizing}
Laurens van~der Maaten and Geoffrey Hinton.
\newblock 2008.
\newblock Visualizing data using t-sne.
\newblock {\em Journal of machine learning research}, 9(Nov):2579--2605.

\bibitem[\protect\citename{Marra \bgroup et al.\egroup
  }2018]{marra2018unsupervised}
Giuseppe Marra, Andrea Zugarini, Stefano Melacci, and Marco Maggini.
\newblock 2018.
\newblock An unsupervised character-aware neural approach to word and context
  representation learning.
\newblock In {\em International Conference on Artificial Neural Networks},
  pages 126--136. Springer.

\bibitem[\protect\citename{Mikolov \bgroup et al.\egroup
  }2010]{mikolov2010recurrent}
Tom{\'a}{\v{s}} Mikolov, Martin Karafi{\'a}t, Luk{\'a}{\v{s}} Burget, Jan
  {\v{C}}ernock{\`y}, and Sanjeev Khudanpur.
\newblock 2010.
\newblock Recurrent neural network based language model.
\newblock In {\em Eleventh annual conference of the international speech
  communication association}.

\bibitem[\protect\citename{Zampieri and Becker}2013]{zampieri2013colonia}
Marcos Zampieri and Martin Becker.
\newblock 2013.
\newblock Colonia: Corpus of historical portuguese.
\newblock {\em ZSM Studien, Special Volume on Non-Standard Data Sources in
  Corpus-Based Research}, 5:69--76.

\bibitem[\protect\citename{Zampieri and Nakov}2020]{zampieri2020similar}
Marcos Zampieri and Preslav Nakov.
\newblock 2020.
\newblock {\em Similar Languages, Varieties, and Dialects: A Computational
  Perspective}.
\newblock Cambridge University Press.

\bibitem[\protect\citename{Zhang and Lapata}2014]{zhang2014chinese}
Xingxing Zhang and Mirella Lapata.
\newblock 2014.
\newblock Chinese poetry generation with recurrent neural networks.
\newblock In {\em Proceedings of the 2014 Conference on Empirical Methods in
  Natural Language Processing (EMNLP)}, pages 670--680.

\bibitem[\protect\citename{Zugarini \bgroup et al.\egroup
  }2019]{zugarini2019neural}
Andrea Zugarini, Stefano Melacci, and Marco Maggini.
\newblock 2019.
\newblock Neural poetry: Learning to generate poems using syllables.
\newblock In {\em International Conference on Artificial Neural Networks},
  pages 313--325. Springer.

\end{thebibliography}

\end{document}